\documentclass[sigconf]{acmart}
\AtBeginDocument{%
  }

\usepackage{wrapfig}

\newcommand{\eg}[0]{e.g.,}
\newcommand{\ie}[0]{i.e.,}

\copyrightyear{2026}
\acmYear{2026}
\setcopyright{cc}
\setcctype{by}
\acmConference[DIS '26]{Designing Interactive Systems Conference}{June 13--17, 2026}{Singapore, Singapore}
\acmBooktitle{Designing Interactive Systems Conference (DIS '26), June 13--17, 2026, Singapore, Singapore}
\acmDOI{10.1145/3800645.3813070}
\acmISBN{979-8-4007-2563-0/2026/06}

\usepackage{pifont}
\usepackage{makecell}
\usepackage{graphicx}   
\usepackage{multirow}
\usepackage{colortbl}
\usepackage{colortbl}
\usepackage{enumitem}

\newcommand{\vcat}[1]{%
  \rotatebox{90}{\parbox{2.2cm}{\centering\textbf{#1}}}%
}
\definecolor{rowgray}{gray}{0.93}
\newcommand{\g}[1]{\cellcolor{rowgray}#1}
\newcommand{\bl}[0]{\color{black}}
\newcommand{\bk}[0]{\color{black}}

\begin{document}

\title{Designing for Robot Wranglers: A Synthesis of Literature and Practice}


\author{David Porfirio}
\affiliation{
  \institution{Computer Science Department}
  \institution{George Mason University}
  \city{Fairfax, VA}
  \country{USA}
}
\email{dporfiri@gmu.edu}

\author{Ian McDermott}
\affiliation{
  \institution{Department of Computer Science/Department of Art}
  \institution{University of Maryland}
  \city{College Park, MA}
  \country{USA}
}
\email{imcdermo@umd.edu}

\author{Hsin-Mei Chen}
\affiliation{
  \institution{Houston Methodist}
  \city{Houston, TX}
  \country{USA}
}
\email{hchen5@houstonmethodist.org}

\author{Satoru Satake}
\affiliation{
  \institution{DIL}
  \institution{ATR}
  \city{Souraku-gun, Kyoto}
  \country{Japan}
}
\email{satoru@atr.jp}

\author{Takayuki Kanda}
\affiliation{
  \institution{Kyoto University}
  \city{Kyoto}
  \country{Japan}
}
\email{kanda@i.kyoto-u.ac.jp}

\author{Thomas D. LaToza}
\affiliation{
  \institution{Department of Computer Science}
  \institution{George Mason University}
  \city{Fairfax, VA}
  \country{USA}
}
\email{tlatoza@gmu.edu}

\renewcommand{\shortauthors}{Porfirio et al.}
\newcommand{\cmark}{\ding{51}}
\newcolumntype{Y}{>{\raggedright\arraybackslash}X}
\newcommand{\david}[1]{\bl{}\textbf{David: }#1\color{black}}
\newcommand{\qt}[1]{\textit{``#1''}}

\begin{abstract}


Robots are increasingly \bl{}present in human spaces\bk{}, such as for conducting deliveries in hospitals, interacting with visitors at museums, and stocking items in warehouses. To ensure the seamless integration of robots into these spaces, a new role in human-robot interaction is emerging---the \textit{robot wrangler}, namely an individual who is responsible for setting up, overseeing, and troubleshooting the robot. To understand the needs of this stakeholder, we conducted a scoping review \bl{}that uncovered a typology of robot wrangling across the research literature, and \color{black} discovered that wrangling \bl{} is an umbrella term that collapses a highly complex and heterogeneous space of activities, often rendering this labor difficult to characterize and support. To further clarify and understand robot wrangling, we then reflected on our own firsthand and imagined experiences as robot wranglers within our own respective domains. Guided by the scoping review and our reflections, we devise a series of design implications for supporting wranglers directly as individuals and as members of a wider service ecology\color{black}.


\end{abstract}




\maketitle

\section{Introduction}

\bl{}Looking beyond the end user, robots are entangled in a distributed ecology of stakeholders, including the \textit{bystanders} who observe and are affected by their presence, the \textit{operators} and \textit{wizards} that monitor and control these robots behind the scenes, and the \textit{developers} and \textit{engineers} who provide technical support. Many of these stakeholders receive sustained attention from the design community, ranging from participatory design for integrating robots into care homes \cite{stegner2022designing} to teleoperation interfaces for remotely controlling robots in the home \cite{cabrera2021exploration}. \color{black}

\bl{}Another stakeholder, the \textit{robot wrangler}, has recently garnered increasing focus in both popular media and in the research community. \color{black}
Defined by Björling and Riek \cite{bjorling2022designing} as ``a human who is responsible for ensuring the robot continues to function, stay charged, and overall ensures a smooth and successful interaction'', many companies now seek to hire robot wranglers in order to ensure that their robot fleets run smoothly. 
As early as 2018, the BBC reported that several technology companies---Aethon,\footnote{\url{https://aethon.com/}} Robby Technologies,\footnote{\url{https://www.ycombinator.com/companies/robby-technologies}} Marble (later acquired by Caterpillar),\footnote{\url{https://promusventures.com/project/marble-robot/}} and Starship\footnote{\url{https://www.starship.xyz/}}---were actively hiring individuals into the wrangling role \cite{sherman2018wrangler}. 
Companies that employ wranglers have enjoyed substantial media attention, such as the Wall Street Journal's reporting on the necessity of robot wranglers in troubleshooting warehouse robots \cite{young2024wrangler}, and CBS News' reporting on the role of wranglers in managing multiple robots at a time \cite{cooper2021wrangler}. \bl{} Simultaneously, the prevalence of the term ``wrangler'' has exploded in its use within robotics literature. \color{black}

However, when examining the term in isolation, without importing domain-specific or colloquial interpretations, the boundaries of what constitutes robot wrangling are fuzzy.
This is owed, in part, to the fact that most research literature mentions ``robot wrangling'' only in passing, rather than as a central part of any research question \citep[\eg{}][]{raibert1986running}.
As a result, when wrangling is mentioned in the literature, whether its primary function involves setting up, maintaining, or troubleshooting robots is unclear.
The individuals who conduct wrangling are also underreported.
Some research positions graduate students who ensure that experiments run smoothly behind the scenes as robot wranglers \cite{lim2023feeding}.
Absent professional support, other literature positions the end users of robots themselves as wranglers \cite{bjorling2022designing}.
\bl{}Overall, wrangling is exceptionally nebulous in its description, making it difficult to understand what aspects of wrangling are distinct from other stakeholders, how to recognize wrangling perspectives, and how to support them. 
\color{black}


\bl{}This paper therefore comprehensively characterizes robot wrangling, including its novel parts and its relationship with other existing, less nebulous, and more explicitly designed-for stakeholders in human-robot interaction (HRI). Given this characterization, we offer implications for supporting the undersupported aspects. \color{black} To characterize wrangling, we begin by conducting a scoping review of ``wrangling'' in the context of robotics in order to understand how the term is perceived in the research literature. The specific research question that guides our review is: \textit{what is the current understanding of ``robot wrangling'' in the research literature, specifically \textbf{who} are wranglers, \textbf{when} and \textbf{where} does wrangling occur, \textbf{what} are wranglers’ responsibilities, and \textbf{why} is wrangling needed?} A key finding of the literature review is the sheer breadth of wrangling accounts, with wrangling resisting definition in terms of any single activity or type of individual. 

\bl{}We then reflect on our own lived and imagined experiences as robot wranglers while recognizing how our social and professional identities as roboticists and software engineers affect this understanding. Our reflection, combined with our scoping review, revealed several themes, which we use to compose a series of vignettes as analytical vehicles that foreground the complexities of wrangling. To provide insight for addressing the tradeoffs and tensions that surface within our themes, we conclude by providing a set of design implications for supporting wrangling in the future. \bk{} Our specific contributions in this paper include:

\begin{itemize}
    \item \textit{Review}---A scoping review that results in a typology and analysis of robot wrangling in the literature.
    \item \textit{Analytical Vignettes}---\bl{}Lived and imagined \color{black} experiences that \bl{}expand upon, clarify, and contextualize \color{black} the concepts uncovered in the scoping review.
    \item \textit{Design}---\bl{}Implications for designers, researchers, and practitioners for supporting wrangling\color{black}, based on our typology and lived and imagined experiences.
\end{itemize}


\section{Background}

This research is fundamentally about the emergence, or perhaps resurgence, of a \bl{}nebulous, though oft-mentioned stakeholder \color{black} in human-robot interaction---the \textit{robot wrangler}. \bl{}We therefore seek to situate our investigation within ongoing DIS conversations on service ecologies, the stakeholders therein, and how service design can support stakeholders within these ecologies. 

\subsection{A Service Ecological Framing of HRI}
The HRI community disproportionately represents 
\bl{}what \citet{bjorling2022designing} refer to as the \textit{primary user},\bk{} encompassing people who are directly affected by the robot's interactions with end users, including the end users themselves, in addition to family and friends. \bl{}Primary users are given substantial research attention for communicating with \citep[\eg{}][]{cruz2025poder, kubota2020jessie, lee2024rex, pham2025gesture, stegner2024understanding}, controlling \citep[\eg{}][]{hagenow2024system, rakita2019shared, alves2022flexi}, and being assisted by robots \citep[\eg{}][]{stegner2022designing, bhattacharjee2020more, reeder2010breakbot}\bk{}.  
%
%
%
%
%
The \textit{secondary} user includes bystanders (those who observe or are co-present with the robot) and designers themselves\bl{}, including those who design experiences or create artifacts for primary users\color{black}. \textit{Tertiary} users, by contrast, include individuals who are indirectly affected by the robot's interactions with primary users, such as the administrators of \bl{}large organizations that receive robot services. 
We observe that in classifying these stakeholders as \textit{secondary} and \textit{tertiary} users, the primary user outshines other stakeholders by remaining at the center of design focus. The needs of other stakeholders are thus framed through the user, exacerbating the invisibility of other crucial support roles.

Reframing HRI beyond the user, \citet{dobrosovestnova2025beyond} provide a service ecology framework that
distinguishes service providers and receivers from service \textit{upkeepers}. Upkeep, in particular, presents several organizationally and ethically complex design challenges. Contrary to the rosy vision of robotic assistance, \citet{tornberg2021investigating} shatters the expectation that robotics systems deployed in the real world are ``plug and play.'' Upkeep therefore highlights an ethical tension in HRI that is recently gaining attention\color{black}---while often being viewed as a source of new labor opportunities in a rapidly evolving economical landscape, upkeep has thus far added to the existing responsibilities of individuals whose work is conducted thanklessly behind the scenes \cite{fox2023patchwork, tornberg2021investigating}. \bl{} In addition to the upkeepers are the incidentally co-present people who are passively and actively affected by the robot's presence \cite{friedrich2025evaluating, friedrichevaluation}, including both bystanders \cite{onnasch2021taxonomy} and the \textit{helpers} that choose to assist the robot when necessary \cite{dobrosovestnova2025beyond}. \citet{fallatah2020would} finds that bystanders already inherently desire to help robots, and \citet{yu2024encouraging} shows how this can be made a playful experience. While playfulness has been shown to improve bystander willingness to help robots, the gamification of upkeep can lead to worker and bystander exploitation \cite{yu2024encouraging, dubbell2015invisible}. Overall, it is important to understand if wrangling is prone to similar tensions, especially given its rapid rise in both the research community and popular media.

\subsection{Practicing Service Design in HCI and HRI}

Supporting the full ecology of stakeholders in HRI requires moving beyond purely technosolutionist approaches. Originally intended for designing the interactions between stakeholders in business operations and management, \textit{service design} is gaining traction in human-computer interaction \cite{lee2022hci}, loosely defined as the backstage work, collaborative networks, ecosystem and infrastructure, and user (or classically, ``customer'') experience associated with designing technology. Recent work in the DIS community underscores the increasing overlap in user experience (UX) and service design, where UX must increasingly consider the interaction of one or multiple stakeholders with a technology service rather than a standalone product \cite{roto2021service}. This overlap is particularly apparent in several user-centric design instances: \citet{davoli2014materializing} probed the internals of a highly opaque delivery service in order to uncover opportunities for community-driven interventions to the delivery status quo in rural areas, while \citet{meurer2020wizard} similarly investigates passenger experiences with robotic taxis. \citet{kim2018sketchstudio} interestingly operationalize a service design concept, \textit{user journey maps}, within \textit{SketchStudio} by enabling service designers to map out complex interactions between a technology and its stakeholders. 

Another tool in service designers' toolkit is \textit{service blueprinting}, in which the designer maps out the customer-facing, frontstage, and backstage interactions to understand the flow of interactions that must occur therein \cite{lee2009designing}. Service blueprinting has begun to be adopted by the HCI community in cases such as rescue services \cite{luhtala2018proactive}, clinical work \cite{yildirim2024investigating}, and deceptive or manipulative patterns in digital services \cite{gray2025getting}, but has only sparingly been applied to HRI \citep[\eg{}][]{elbeleidy2024preliminary, lee2009designing}. In investigating emerging stakeholders such as robot wranglers, understanding how these wranglers are situated within the HRI ecosystem can benefit from tools such as these.
\bk{}







\section{Scoping Review}\label{sec:review}

\begin{figure}
    \centering
    \includegraphics[width=1\columnwidth]{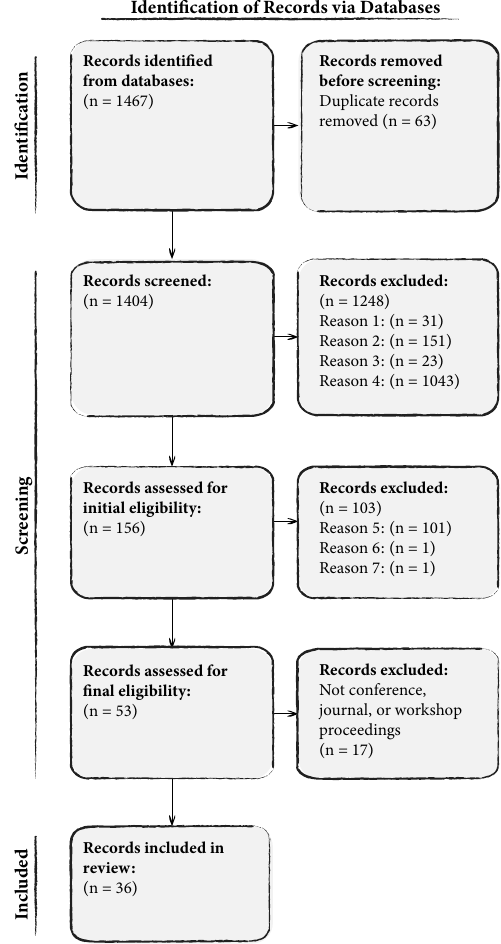}
    \caption{A PRISMA flow diagram that illustrates our review process. \bl{}The diagram is read from top to bottom, with the left column indicating the number (\textit{n}) of records retained at each point during the review phase, and the right column indicating the reasons for record removal.\color{black}}
    \Description{This PRISMA diagram contains five rows of information about how works were excluded from the review. The first row states that 1467 records were identified from all databases included in the search, and 63 were subsequently removed as duplicates. The second row states that 1404 works resulted and were screened after the previous elimination, and 1248 were excluded due to reasons one through four in the text. The third row states that 156 works resulted and were assessed for initial eligibility resulting from the previous elimination, and that another 103 were excluded due to reasons five through seven in the text. The fourth row states that 53 records resulted and were assessed for final eligibility resulting from the previous exclusion, and that 17 were excluded due to not being a conference journal, or workshop article. The last row states that 36 records were kept for review.}
    \label{fig:prisma}
\end{figure}

We begin with a multi-phase scoping review to understand the research community's existing view of robot wrangling. Our reporting of this review is guided by the standards of the Preferred Reporting Items for Systematic Review and Meta-Analysis (PRISMA) \cite{page2021prisma}. Our primary inclusion criteria for our review is that ``wrangling'' as it pertains to robots is described in \textit{any} level of detail in \textit{any} section of the works included, even if only mentioned in the related work or discussion sections, and even if not related to the primary contributions of the paper. These inclusion criteria are informed by our desire to understand the research community's view of wrangling, not necessarily contributions made therein.

Figure \ref{fig:prisma} illustrates our review process. In what follows, we describe our \textit{identification} and \textit{screening} steps of the review.

\subsection{Identification}

We conducted our search on December 22, 2025. First, we drew from four databases that publish work in robotics: \textit{ACM}, \textit{IEEE}, \textit{Springer Nature Link}, and \textit{Science Direct}. Our search encompassed the full text and metadata of all works returned from the search query: \texttt{robot* AND wrangl*}. Following \citet{ajaykumar2021eup}, we then included a fifth source to include any remaining works: the first 100 works that appear in \textit{Google Scholar} using the same search query.

Our search yielded a total of 1,467 works, with 246 being returned from \textit{ACM}, 313 being returned from \textit{IEEE}, 576 being returned from \textit{Springer Nature Link}, 232 being returned from \textit{Science Direct}, and 100 being returned from \textit{Google Scholar}. We then removed 63 duplicates for a total of 1,404 works.

\subsection{Screening}

Our first screening step included the following exclusion criteria:

\begin{itemize}[leftmargin=1.5em]
  \item \textbf{Reason 1: nonexistence}---we excluded nonexistent work, including dead links in \textit{Google Scholar}, retracted papers, and book chapters consisting only of indices.
  \item \textbf{Reason 2: frontmatter}---we excluded conference proceedings frontmatter, which was often found in the ACM Digital Library.
  \item \textbf{Reason 3: non-English}---we excluded works that are not written in English.
  \item \textbf{Reason 4: irrelevant topic}---if none of the previous exclusion criteria applied, the first author read the title and abstract of the work to determine if its topic related to robotics or human-robot interaction. The paper was excluded if not related. For book chapters, editorials, and other works that do not have abstracts, the first author read the first few paragraphs of the work.
\end{itemize}

If the title and abstract indicated that the work may be relevant to robotics or human-robot interaction, we then assessed each record for initial eligibility by applying the following exclusion criteria:

\begin{itemize}[leftmargin=1.5em]
  \item \textbf{Reason 5: no ``robot'' wrangling}---the first author searched for the keyword ``robot'' and prefix ``wrangl'' in the body of text. If ``wrangl'' was used separately from describing any activity conducted with the robot, this exclusion criteria applied. For example, we found several robot-relevant works in which the term ``wrangling'' is used in the context of data wrangling or robot grasping.
  \item \textbf{Reason 6: robot wrangling without any context}---if the work includes a mention of robot wrangling without any context whatsoever, the work was excluded. This exclusion criteria applied to only one work: a Ph.D. thesis in which ``robot wrangler'' is mentioned only in the acknowledgments.
  \item \textbf{Reason 7: paper inaccessible to authors}---if no authors' institutions have access to the work, the work was excluded. This exclusion criteria applied to only one work.
\end{itemize}

\begin{figure}
    \centering
    \includegraphics[width=\columnwidth]{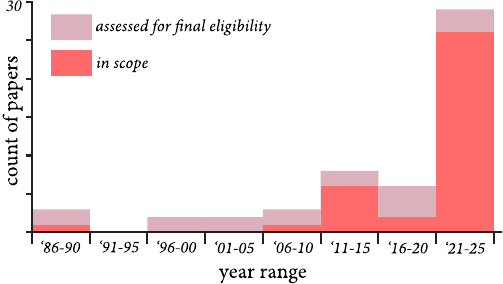}
    \caption{The prevalence of papers that mention, discuss, or are about robot wrangling between the years 1986 (the earliest record we could find) and 2025. The lightly-colored bars represent records assessed for final eligibility, and the darker-colored bars represent in-scope records included in the review. The higher prevalence of the term in the last five years can be partially attributed to a keynote given by Leila Takayama in 2022 \cite{takayama2022putting}, the abstract for which is in the group assessed for final eligibility, and which several of our in-scope papers cite.}
    \Description{A stacked bar chart showing the number of papers on robot wrangling across eight year ranges from 1986–1990 through 2021–2025. The x-axis lists year ranges; the y-axis shows count of papers (0–30). Each bar has two segments: a lighter segment for papers assessed for final eligibility and a darker segment for papers included in scope. Very few papers appear before 2005, with small counts in 1986–1990 and minimal or none in 1991–2005. Counts increase modestly from 2006–2015, grow further in 2016–2020, and rise sharply in 2021–2025, which has by far the highest total (nearly 30 assessed, with most in scope). The chart highlights a strong recent increase in publications using the term “robot wrangling.}
    \label{fig:prevalence}
\end{figure}

This phase of the review returned a total of 53 works---28 conference articles, 7 journal articles, 1 workshop article, 5 IEEE magazine articles, 2 textbook chapters, 1 keynote talk, 6 theses, 1 course project, 1 novel of fiction, and 1 work of unknown nature. In including all of these works in this phase, we are able to characterize the historical use of the term ``wrangling'' within robotics over time. Figure \ref{fig:prevalence} depicts the historical prevalence of the term.

In order to answer our research question, the final phase of our screening process narrowed the selection of papers from the previous page to \textit{individual works of research}---conference, journal, and workshop articles, including companion proceedings. Applying this refined inclusion criteria to our selection of works, our final count of papers that are in-scope for this review is 36.

\subsection{Analysis of In-Scope Papers}

Our analysis of in-scope papers focused on developing a typology of robot wrangling as described by existing literature and informed by the experiences of robot wranglers and stakeholders of technology deployed in the workforce. To begin this process, the first author, who has firsthand robot wrangling experience, took informal notes on the description of wrangling within all papers, namely \textit{who} the wranglers were, \textit{what} wrangler duties entailed, \textit{where} and \textit{when} wrangling occurred, and \textit{why} wrangling was necessary. From these notes, the first author developed an initial \bl{}codebook\color{black}, then iteratively refined \bl{}the codebook \color{black} by discussing it with (1) the second author, who also has firsthand robot wrangling experience; (2) the third author, who is part of an organization at the forefront of \bl{}technology integration in healthcare\color{black};
and (3) the last author, who is an expert in software engineering, a critically relevant field to robot wrangling.

\bl{}Using the codebook\color{black}, the first and final author then \bl{}independently conducted an initial round of inductive (open) coding \color{black} on nine papers (25\% of the in-scope corpus) based on these categories. They met twice to \bl{}compare codes, resolve discrepancies, and iteratively refine the category set. Disagreements most often arose when papers implicitly referred to practices deemed relevant to our review without drawing an explicit link to wrangling\color{black}. As an outcome of this process, the authors established two coding principles: (1) each paper would be analyzed in isolation, without influence from other papers’ interpretations of “wrangling,” and (2) only \textit{explicit} links between the term “wrangling” and described activities would be coded. For example, a paper focused on teleoperation may be interpreted---based on other literature---as involving wrangling; however, if the paper does not explicitly connect teleoperation to ``wrangling,'' it was not coded as such. Using the finalized category set and coding criteria, the first author then coded the remaining papers. Table \ref{tab:categorization} presents the resulting typology.



\renewcommand{\arraystretch}{1.15}

\small
\begin{table*}[t]
\small
\centering
\caption{Papers categorized by how wrangling is described. This categorization is based on direct links between the term ``wrangling'' and the activities described in the paper.}
\Description{A table categorizing papers on robot wrangling across four dimensions: who performs wrangling (e.g., designated, undesignated, trained, untrained, team, individual), type of intervention (physical, non-physical), purpose (e.g., maintain, protect, orchestrate, intervene, teleoperate, explain, troubleshoot, tidy), where it occurs (e.g., co-located, remote, public, non-public, frontstage, backstage), and when it occurs (e.g., synchronous, asynchronous, spontaneous, planned, continuous or triggered monitoring), with brief definitions and representative citations for each category.}
\label{tab:categorization}

\begin{tabular}{%
    >{\centering\arraybackslash}p{16pt}
    >{\arraybackslash}p{50pt}
    >{\arraybackslash}p{276pt}
    >{\arraybackslash}p{124pt}
}
\toprule
&
\textbf{Category} &
\textbf{Description (Wrangling is described as...)} &
\textbf{Main ref.} \\
\midrule

\multirow{3}{*}{\vcat{Who}}
& designated
& involving individuals who have been designated for the wrangling role.
& \cite{pelikan2025people, 
lim2023feeding, 
lee2025minding, 
ore2015autonomous, 
benford2025tangles, 
riek2025future, 
boaventura2012dynamic, 
hunt2025children, 
schneiders2024designing, 
benford2025somatic, 
verzijlenberg2010swimming, 
silvis2022technical, 
stoddarduser, 
raibert1986running, 
gamboa2023wisp, 
bjorling2022designing, 
ngo2024dancing, 
benford2024artists, 
speers2013diver, 
gamboa2025we, 
schneiders2023tas, 
colett2012artificial, 
francis2025principles, 
benedict2019application, 
pelikan2024encountering, 
speers2011monitoring, 
chiou2022towards, 
ghosh2025envisioning, 
anderson2024integrated, 
martin2022towards, 
edge2020design, 
doniec2013robust, 
paton20242023, 
wilson2023exploring, 
benford2024charting, 
benford2025charting} \\ 

&
\g{undesignated}
& \g{involving individuals who have assumed wrangling duties without being designated for the wrangling role.}
&  \g{\cite{lee2025minding, bjorling2022designing}} \\ 

& trained
& involving individuals who have received training for the wrangling role.
& \cite{pelikan2025people, 
lim2023feeding, 
benford2025tangles, 
gamboa2025we, 
francis2025principles, 
chiou2022towards, 
martin2022towards, 
doniec2013robust, 
paton20242023, 
wilson2023exploring} \\ 

&
\g{untrained}
& \g{involving individuals who have \textbf{\textit{not}} received training for the wrangling role.}
& \g{\cite{hunt2025children, 
silvis2022technical, 
benedict2019application, 
anderson2024integrated}}  \\ 

&
team
& involving a team of multiple wranglers.
& \cite{pelikan2025people, 
lim2023feeding, 
boaventura2012dynamic, 
verzijlenberg2010swimming, 
speers2013diver, 
paton20242023} \\ 

&
\g{individual}
& \g{involving an individual who is the sole wrangler, even if the wrangler is part of a larger team.}
& \g{\cite{ore2015autonomous, 
hunt2025children, 
benford2025somatic, 
stoddarduser, 
gamboa2023wisp, 
ngo2024dancing, 
speers2013diver, 
gamboa2025we, 
schneiders2023tas, 
colett2012artificial, 
benedict2019application, 
anderson2024integrated, 
edge2020design, 
wilson2023exploring}} \\ 

\midrule

\multirow{3}{*}{\vcat{Interv.}}
& Physical
& involving physical interaction with the robot.
& \cite{pelikan2025people, 
lim2023feeding, 
lee2025minding, 
ore2015autonomous, 
benford2025tangles, 
boaventura2012dynamic, 
schneiders2024designing, 
benford2025somatic, 
verzijlenberg2010swimming, 
silvis2022technical, 
raibert1986running, 
speers2013diver, 
colett2012artificial, 
speers2011monitoring, 
doniec2013robust, 
paton20242023, 
wilson2023exploring, 
benford2025charting} \\ 

&
\g{Non-physical}
& \g{involving non-physical interaction with the robot.}
& \g{\cite{pelikan2025people, 
ore2015autonomous, 
stoddarduser, 
ngo2024dancing, 
chiou2022towards, 
benford2025charting}} \\ 

\\

\midrule

\multirow{3}{*}{\vcat{Purpose}}
& Maintain
& routine maintenance tasks before or between sessions such as setting up, resetting, charging, or re-parameterizing the robot.
& \cite{pelikan2025people, 
lim2023feeding, 
lee2025minding, 
riek2025future, 
schneiders2024designing, 
benford2025somatic, 
stoddarduser, 
bjorling2022designing, 
speers2013diver, 
pelikan2024encountering, 
paton20242023} \\ 

&
\g{Protect}
& \g{ensuring the safety of the robot or individuals who interact with the robot.}
& \g{\cite{pelikan2025people, 
ore2015autonomous, 
benford2025tangles, 
boaventura2012dynamic, 
hunt2025children, 
verzijlenberg2010swimming, 
raibert1986running, 
schneiders2023tas, 
colett2012artificial, 
ghosh2025envisioning, 
doniec2013robust, 
benford2025charting}}  \\ 

&
Orchestrate
& providing guidance during the normal operations to another agent involved in the task or interaction, be it the (autonomous) robot itself, another team member, or an individual who interacts with the robot.
& \cite{pelikan2025people, 
lee2025minding, 
benford2025tangles, 
riek2025future, 
stoddarduser, 
ngo2024dancing, 
wilson2023exploring, 
benford2025charting} \\ 

& \g{Intervene}
& \g{adjusting the action of an agent, be it the (autonomous) robot itself, another team member, or an individual who interacts with the robot involved in the task or interaction at the time that the action occurs.}
& \g{\cite{pelikan2025people, 
lee2025minding, 
benford2025tangles, 
boaventura2012dynamic, 
schneiders2024designing, 
benford2024artists, 
colett2012artificial, 
benford2025charting}}  \\ 

&
Teleoperate
& directly controlling the behaviors of the (non-autonomous) robot.
& \cite{pelikan2025people, 
ore2015autonomous, 
riek2025future, 
stoddarduser, 
benford2024artists, 
gamboa2025we, 
chiou2022towards, 
benford2025charting} \\ 

&
\g{Explain}
& \g{explaining any aspect of the robot to other stakeholders.}
& \g{\cite{pelikan2025people, 
benford2025tangles, 
benford2025somatic, 
gamboa2023wisp, 
ngo2024dancing}}  \\ 
&
Troubleshoot
& diagnosing and fixing defects that occur related to the robot, its task, or interaction.
& \cite{pelikan2025people, 
lim2023feeding, 
riek2025future, 
silvis2022technical} \\ 

&
\g{Tidy}
& \g{cleaning up after the robot or after interactions that occur with the robot.}
& \g{\cite{lim2023feeding, 
benford2025tangles, 
schneiders2024designing}}  \\ 

\midrule

\multirow{3}{*}{\vcat{Where}}
& Co-located
& being physically co-located with the robot during any wrangling activity.
& \cite{pelikan2025people, 
lim2023feeding, 
lee2025minding, 
ore2015autonomous, 
benford2025tangles, 
boaventura2012dynamic, 
hunt2025children, 
schneiders2024designing, 
benford2025somatic, 
verzijlenberg2010swimming, 
silvis2022technical, 
raibert1986running, 
gamboa2023wisp, 
speers2013diver, 
colett2012artificial, 
speers2011monitoring, 
ghosh2025envisioning, 
anderson2024integrated, 
edge2020design, 
doniec2013robust, 
paton20242023, 
wilson2023exploring, 
benford2025charting} \\ 

&
\g{Remote}
& \g{\textbf{\textit{not}} being physically co-located with the robot during a wrangling activity.}
& \g{\cite{ore2015autonomous, 
stoddarduser, 
ngo2024dancing, 
gamboa2025we, 
benford2025charting}}  \\ 

&
public
& involving a robot that exists in public view. 
& \cite{pelikan2025people, 
lee2025minding, 
ore2015autonomous, 
benford2025tangles, 
ngo2024dancing, 
gamboa2025we, 
pelikan2024encountering, 
benford2024charting, 
benford2025charting} \\ 

& \g{Non-public}
& \g{involving a robot that does not exist in public view.}
& \g{\cite{boaventura2012dynamic, 
hunt2025children, 
benford2025somatic, 
verzijlenberg2010swimming, 
raibert1986running, 
speers2013diver, 
colett2012artificial, 
speers2011monitoring, 
edge2020design, 
doniec2013robust, 
paton20242023}} \\ 

&
Frontstage
& involving activities that occur in public view, namely activities that can be viewed by undesignated bystanders 
& \cite{pelikan2025people, 
benford2025tangles, 
ngo2024dancing, 
benford2025charting} \\ 

&
\g{Backstage}
& \g{involving activities that do \textbf{\textit{not}} occur in public view with activities that can be viewed by undesignated bystanders, even though the robot exists in public.}
& \g{\cite{pelikan2025people, 
riek2025future, 
benford2024artists, 
gamboa2025we, 
pelikan2024encountering, 
benford2025charting}}  \\ 

\midrule

\multirow{3}{*}{\vcat{When}}
& Synchronous
& occurring at the same time as when the robot is deployed.
& \cite{
pelikan2025people, 
lee2025minding, 
ore2015autonomous, 
benford2025tangles, 
riek2025future, 
boaventura2012dynamic, 
hunt2025children, 
schneiders2024designing, 
benford2025somatic, 
verzijlenberg2010swimming, 
silvis2022technical, 
stoddarduser, 
raibert1986running, 
gamboa2023wisp, 
benford2024artists, 
speers2013diver, 
gamboa2025we, 
colett2012artificial, 
speers2011monitoring, 
edge2020design, 
doniec2013robust, 
paton20242023, 
wilson2023exploring, 
benford2025charting} \\ 

&
\g{Asynchronous}
& \g{occurring at a different time than when the robot is deployed.}
& \g{\cite{riek2025future, 
ngo2024dancing}} \\ 

&
Spontaneous
& involving little to no preparation for the procedures involved.
& \cite{pelikan2025people, 
lim2023feeding, 
lee2025minding, 
benford2024artists, 
gamboa2025we, 
benford2025charting}  \\ 

&
\g{Planned}
& \g{involving substantial preparation for the procedures involved.}
& \g{\cite{pelikan2025people, 
lim2023feeding, 
ore2015autonomous, 
ngo2024dancing, 
gamboa2025we, 
paton20242023, 
benford2025charting}} \\ 

&
Cont. monitor
& involving monitoring the robot continuously or within pre-determined intervals.
& \cite{pelikan2025people, 
lee2025minding, 
ore2015autonomous, 
benford2025tangles, 
riek2025future, 
boaventura2012dynamic, 
verzijlenberg2010swimming, 
stoddarduser, 
raibert1986running, 
benford2024artists, 
speers2013diver, 
gamboa2025we, 
chiou2022towards, 
edge2020design, 
wilson2023exploring, 
benford2025charting}  \\ 

&
\g{Trig. monitor}
& \g{involving specific events that trigger monitoring, which would not otherwise occur.}
& \g{\cite{lee2025minding, 
riek2025future, 
benford2025charting}} \\ 

\bottomrule
\end{tabular}
\end{table*}
\normalsize

\subsection{Findings}

We discuss our findings below, referring in this section to the typology from Table \ref{tab:categorization} in \textit{italics}. We then highlight our main design implication that emerged from our review---\bl{}\textit{wrangling serves as a metaphorical framing through which researchers interpret and describe a wide variety of activities and roles.}\color{black}

\paragraph{\textbf{What} does wrangling entail and \textbf{when} and \textbf{where} does it occur?}
Our review highlights that wrangling is comprised of many different kinds of activities that are performed in various different ways, places, and times. This is especially evident in the \textit{physicality} of wrangling. More papers describe wrangling as \textit{physical} over \textit{non-physical} activity, for which there appears to be a historical basis in holding a physical tether attached to a robot during laboratory tests in order to ensure the safety of the research equipment \cite{raibert1986running}. This ``tethering'' description of robot wrangling is common to several other works \cite{colett2012artificial, boaventura2012dynamic, ore2015autonomous}, usually as a \textit{continuous}, \textit{synchronous} form of robot wrangling, though not always necessarily just for safety \cite{doniec2013robust, speers2011monitoring, speers2013diver, verzijlenberg2010swimming, edge2020design}. 
Recently, physical wrangling has embodied a different set of activities, sometimes referring to \textit{spontaneous} interventions that wranglers must practice for autonomous or teleoperated robots in public \cite{lee2025minding, pelikan2025people}, routine \textit{maintenance} that must occur with these robots \cite{lim2023feeding, schneiders2024designing}, and for helping facilitate somaesthetic interaction \cite{benford2025tangles, benford2025somatic, martin2022towards, wilson2023exploring}, to name a few examples. In most of these examples, the wrangler is described as being \textit{co-located} with the robot.

Not all wrangling involves physical interaction. We identified several papers that, beginning in 2015, describe definitively \textit{non-physical} wrangling activity. A subset of these works equate ``wrangling'' the robot to directly controlling, or ``operating'' the robot from a distance \cite{pelikan2025people, stoddarduser, ore2015autonomous}. \citet{gamboa2025we} further argues that wrangling is a direct precursor to skilled teleoperation, namely \qt{being and becoming a robot.} This is especially interesting in light of other papers, by contrast, drawing a distinction between wrangling and operating \cite{benford2025tangles, hunt2025children, doniec2013robust, speers2011monitoring, speers2013diver, verzijlenberg2010swimming, edge2020design}. 

It is then unsurprising that wrangling is described as consisting of so many different types of activities, the most common being \textit{maintaining} the robot and ensuring the safety of (\textit{protecting}) the robot or people around it, with \textit{orchestrating} or \textit{intervening} in the robot's (possibly autonomous) operations and \textit{teleoperating} the robot following closely behind. Still, we noted a few surprises. \textit{Explaining}, in particular, or discussing the robot with bystanders or end users and possibly talking through their interactions with the robot, is a somewhat common activity described as wrangling \cite{pelikan2025people, benford2025tangles, benford2025somatic, gamboa2023wisp, ngo2024dancing}. \textit{Tidying} up after the robot also appears in a few papers \cite{lim2023feeding, benford2025tangles, schneiders2024designing}. Rarely do papers explicitly describe wrangling activities as being \textit{asynchronous} (occurring at a different time to when the robot is deployed), making one paper stand out in particular---a ``wrangler'' at a robot art exhibit consisting of a pre-recorded character \cite{ngo2024dancing}.



To our surprise, \textit{troubleshooting} is not dominant in the wrangling literature, captured by only a few papers \cite{pelikan2025people, lim2023feeding, riek2025future, silvis2022technical}. We credit our belief that troubleshooting would play an elevated role in wrangling to our own experiences as wranglers and to popular media that equates wrangling to problem diagnosis and fixing \cite{young2024wrangler}. Related to \textit{troubleshooting}, only a small subset of papers in our review explicitly discuss wrangling as \textit{asynchronous} or \textit{triggered} (occurring offline, outside of the robot's designated activities) \cite{riek2025future, ngo2024dancing, benford2025charting, lee2025minding}. We can draw one of two possible conclusions from this finding---either (1) troubleshooting encompasses only a narrow slice of wrangling activity, or (2) troubleshooting is actually a more common activity than our review suggests and our in-scope papers simply fail to describe this activity. Based on our own experience as wranglers (see Section \S\ref{sec:vignettes}), we believe the latter is more plausible.




\paragraph{\textbf{Who} is the wrangler?} Most in-scope papers describe wranglers as individuals who have  been \textit{designated}, and in many cases \textit{trained}, for the role. We note that the four papers describing \textit{untrained} wranglers pertain to educational contexts, by definition of which the wranglers would be inexperienced \cite{hunt2025children, silvis2022technical, benedict2019application, anderson2024integrated}. Only two papers describing wranglers as being \textit{undesignated} \cite{lee2025minding, bjorling2022designing}, in which bystanders or end users ``fall into'' the role of wrangling.

\citet{riek2025future} and one of the \textit{undesignated} papers---\citet{bjorling2022designing}---highlight an interesting dichotomy. \citet{bjorling2022designing} present a positive view of a future in which the natural progression of robotics involves transferring wrangling duties from designated individuals to undesignated end users. In doing so, end users will have more agency over their robots, rather than this agency being in the hands of third-party individuals. At the same time, \citet{riek2025future} highlight the perils of wrangling as constituting invisible, thankless labor. The invisibility of wrangling is particularly evident in the \textit{frontstage} and \textit{backstage} distinction for wrangling that occurs in public view. It appears to us that these perils would be exacerbated if the wrangling role were to be left undesignated. However, we believe that there is synergy between both arguments---rather than avoiding undesignated wrangling altogether due to the pitfalls of invisible labor, we as designers must support these individuals on the frontstage in order to afford them with the greater agency advocated by \citet{bjorling2022designing}.

Aside from being designated or trained, we also looked at whether wrangling occurred as a \textit{team} or \textit{individually}. Surprisingly, given the many responsibilities that wranglers must juggle, often necessitating \textit{continuous} monitoring of the robot, wrangling is often explicitly portrayed as an individual activity. This both mirrors and contrasts with our own experiences as wranglers (see Section \S\ref{sec:vignettes}).

\begin{figure}[t]
    \centering
    \includegraphics[width=\columnwidth]{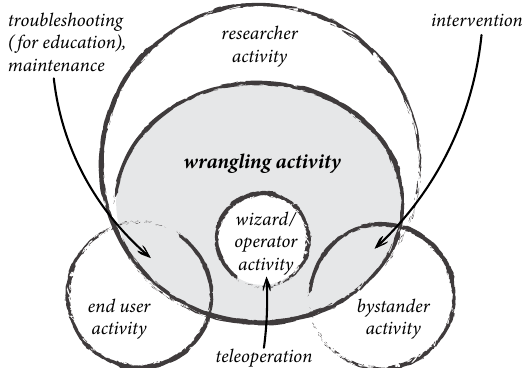}
    \caption{\bl{}How the works in our review position wrangling activities against the activities of a non-exhaustive set of common HRI stakeholders. These works indicate that researchers can be involved in \textit{all} wrangling activities; bystanders when \textit{intervening}; and end users when \textit{troubleshooting} in educational context and \textit{maintaining}. Lastly, these works position operating and wizarding as synonymous with wrangling.\bk{}}
    \Description{A conceptual diagram showing how ``wrangling activity'' relates to different stakeholder roles in human–robot interaction. A large circle labeled ``researcher activity'' contains a smaller circle labeled ``wrangling activity,'' which itself contains a central circle labeled ``wizard/operator activity.'' Two partially overlapping circles labeled ``end user activity'' and ``bystander activity'' intersect with the wrangling circle from below left and right, respectively. Arrows indicate examples of overlap: ``troubleshooting (for education), maintenance'' pointing from end users into wrangling; ``intervention'' pointing from bystanders into wrangling; and ``teleoperation'' pointing upward into the wizard/operator core. The diagram conveys that researchers can engage in all wrangling activities, operators/wizards are central to wrangling, and end users and bystanders participate in more limited, situational ways.}
    \label{fig:stakeholders}
\end{figure}

\bl{}Lastly, the papers in our review discuss several common stakeholders in HRI---the \textit{end user}, the \textit{researcher}, the \textit{wizard} or \textit{operator}, and the \textit{bystander}---and we examine how these roles are situated with respect to wrangling activities. Figure \ref{fig:stakeholders} illustrates our findings. Several papers in our review describe members of the research team themselves as wranglers \citep[\eg{}][]{pelikan2025people, lim2023feeding, paton20242023, gamboa2023wisp}; taken together, these works span the full range of wrangling activities. Accordingly, our review positions the activities that researchers \textit{can} engage in as a \textit{superset} of wrangling activity. 
Conversely, operators (those who directly control robots) and wizards (those who control robots behind-the-scenes) are often described synonymously with wranglers \cite{gamboa2025we, pelikan2025people}; thus, the literature positions wizarding and operating as a \textit{subset} of wrangling activity. For end users, several works in our corpus describe \textit{maintaining} the robot \cite{riek2025future} and \textit{troubleshooting} the robot in educational contexts \cite{hunt2025children, silvis2022technical, benedict2019application, anderson2024integrated} as forms of wrangling. Several other works discuss bystanders as \textit{intervening} when the robot needs assistance \cite{lee2025minding}. We find that this picture of bystander wranglers connects with work outside of our corpus, particularly the notion of wranglers as \textit{helpers} \cite{dobrosovestnova2025beyond, fallatah2020would, yu2024encouraging}.

\textit{Tidying} and \textit{explaining} are overall underrepresented due to the small number of works that capture these activities and the lack of clear non-researcher stakeholders that partake in these roles. Whether wrangling is distinct from other stakeholders in these activities remains to be investigated, though \textit{tidying} after the robot is reminiscent of the invisible \textit{upkeep} and \textit{patchwork} conducted by designated robot support roles \cite{fox2023patchwork, dobrosovestnova2025beyond}. 






\bl{}\paragraph{\textbf{Implication:} wrangling is a metaphor.} Overall, our review shows that wrangling is shaped by researchers’ individual perspectives, collectively encompassing a diverse set of activities carried out by a wide range of people across contexts. Figure \ref{fig:stakeholders} shows how the research community views wrangling as overlapping with several existing roles in HRI---users, wizards, operators, researchers, and bystanders---making it difficult to discern what wrangling \textit{is} and what it \textit{is not}.

To more concretely characterize its nature---whether it manifests as a role, a set of practices, a temporary condition, or an institutional function---we turn to its historical usage for clues. The earliest use of the term ``wrangling'' in our corpus appears in \citet{raibert1986running}, where a wrangler physically stabilizes a robot using a rope for safety, and is echoed in subsequent work \cite{colett2012artificial, boaventura2012dynamic, ore2015autonomous}. We find this historical context enlightening when combined with the definition of the word itself, which Merriam-Webster gives as ``to herd and care for (livestock and especially horses) on the range'' \cite{merriam_webster_wrangler_2026}. Taken together, we argue that wrangling is not a single, well-defined activity, role, place, or function, but a \textit{metaphorical framing} through which researchers interpret and describe the distributed labor involved in achieving reliable human–robot interaction. Through the cattle metaphor, ``wrangling'' becomes a practice, which can be embodied by different roles, for managing unpredictable, semi-autonomous systems that must be guided, contained, corrected, and cared for.

As a metaphor, wrangling is both useful and risky. It helpfully foregrounds the challenges and unpredictability that stakeholders face when dealing with robots. Furthermore, the existence of the term legitimizes the work of certain stakeholders (\eg{} bystanders). On the other hand, wrangling collapses the complexity of the work that it entails to the point of triviality and exceptional vagueness. Crucially, it romanticizes difficult and thankless work by evoking imagery of rugged individualism. While useful as a metaphor, it should be accompanied by reporting practices that more fully capture and make visible the labor it entails. 
\color{black}


\paragraph{Relevant Literature Outside of our Scoping Review:}
As evidenced by works that equate wrangling with \textit{operating} and \textit{wizarding} \cite{gamboa2025we, pelikan2025people}, wrangling is often referred to by several different names. Most notably, \citet{verzijlenberg2010swimming} refers to a ``robot handler'' that bears substantial similarity to wrangling as discussed by many papers in our review. A search for this term outside of our review yields a definition from \citet{woods2004envisioning}: \qt{The robot handler role is responsible for managing the robotic capabilities in situ as a valued resource and points to the knowledge, practice, and interfaces needed to manage the robots in a physical environment.} 
The search for ``robot handler'' yielded a few additional papers \cite{cardenas2019unique, lofaro2015archr}, indicating that \textit{handling} is another way to describe wrangling activities, and more generally, that \textit{wrangling} likely goes by many different names that our scoping review did not pick up.

\bl{}The \textit{undesignated} \cite{bjorling2022designing, lee2025minding} and \textit{backstage} work \cite{pelikan2025people, 
riek2025future, 
benford2024artists, 
gamboa2025we, 
pelikan2024encountering, 
benford2025charting} identified in our review closely parallels another term---\textit{patchwork} \cite{fox2023patchwork}---and more generally, the invisible labor discussed outside of our review \cite{tornberg2021investigating}. However, despite this broader body of work linking wrangling practices to invisible labor, our review surfaces relatively few papers that explicitly frame such activities as \textit{undesignated} or \textit{backstage} work. It is perhaps fitting that this labor remains ``invisible'' within the wrangling literature.\bk{}



\section{\bl{}Lived and Imagined Experiences \color{black} of  Wrangling}\label{sec:vignettes}

\bl{}Whereas our typology reflects a broad, community-level view of robot wrangling, few papers in our review offer a deep account from the perspective of self-identified wranglers. We therefore draw on our own lived and imagined experiences as robot wranglers to help surface this viewpoint. Taken together, our literature review and reflections aim to provide a more complete account of wrangling. 

We engaged in extensive discussions over the course of two months about our own firsthand and imagined experiences as wranglers. The first and final author began by conducting brainstorming sessions with the second and third authors individually over video call, followed by a joint brainstorming session with the fourth and fifth authors. The first and final author met in between brainstorming sessions to compare notes, from which a set of themes emerged, which were refined in subsequent meetings with the other authors. All authors then composed a set of vignettes that foreground these themes in concrete, situated wrangling scenarios.

Each vignette is organized by describing (1) a real-world context and other stakeholders involved; (2) one or more real (and in one case, imagined) scenarios that required wrangling; and (3) a typological characterization of the wrangling that occurred. Below, we describe each vignette, followed by presenting the themes. Table \ref{table:vignettes} depicts the wrangler profiles for each vignette.

\bl{}\paragraph{Positionality:} The authors---five researchers and one practitioner at various academic, healthcare, and private organizations---bring together perspectives from HRI, software engineering, and domain-specific practice, with experience in designing, deploying, and studying robots in public settings. Our conceptualization of ``robot wrangling'' is shaped by this background, which may incline us to view wrangling through an HRI rather than a broader socio-technical design perspective, and through the lens of visible rather than invisible labor. Additionally, our experiences are grounded in specific domains and institutional contexts, which limit the comprehensiveness of our reflection, and privilege technical interventions over organizational and labor-oriented changes. Importantly, our vignettes do not include scenarios such as search and rescue, collaborative robotics, and manufacturing.\color{black}

\renewcommand{\arraystretch}{1.15}
\small
\begin{table*}
\centering
\label{tab:hd_profile}
\begin{tabular}{
    p{80pt}
    p{76pt}
    p{60pt}
    p{76pt}
    p{76pt}
    p{76pt}
}
\toprule
\makecell{\textbf{Vignette}} & \textbf{Who} & \textbf{Intervention} & \textbf{Purpose} & \textbf{Where} & \textbf{When} \\
\midrule
\makecell{\textbf{Help Desk Robot}} & \makecell[l]{team of trained,\\designated individuals} & \makecell[l]{physical} & \makecell[l]{intervention,\\troubleshoot} & \makecell[l]{co-located,\\backstage, public} & \makecell[l]{sync., spontaneous,\\continuous} \\
\midrule
\makecell{\textbf{Fundraising Robot}} & \makecell[l]{team of trained,\\designated individuals} & \makecell[l]{non-physical} & \makecell[l]{maintain,\\teleoperate} & \makecell[l]{co-located,\\backstage, public} & \makecell[l]{synchronous,\\asynchronous, planned} \\

\midrule

\makecell{\textbf{Shopworker Robot}\\Prototyping} & \makecell[l]{team, designated} & \makecell[l]{non-physical} & \makecell[l]{troubleshoot,\\maintain} & \makecell[l]{co-located,\\frontstage, public} & \makecell[l]{synchronous,\\asynchronous, planned} \\

\addlinespace[0.5em]

\makecell{\textbf{Shopworker Robot}\\Deployment} & \makecell[l]{team, designated,\\trained and untrained} & \makecell[l]{non-physical} & \makecell[l]{troubleshoot,\\protect, teleoperate} & \makecell[l]{remote, co-located,\\backstage, public} & \makecell[l]{synchronous, planned,\\asynchronous, triggered}\\

\addlinespace[-0.7em]

\midrule

\makecell{\textbf{Museum Robot}\\Teen Education} & \makecell[l]{trained, designated} & \makecell[l]{non-physical} & \makecell[l]{maintain, explain,\\troubleshoot} & \makecell[l]{back/frontstage\\non-public, co-located} & \makecell[l]{asynchronous, planned\\sync., spontaneous} \\

\addlinespace[0.5em]

\makecell{\textbf{Museum Robot}\\Affordance Mismatch} & \makecell[l]{untrained, designated} & \makecell[l]{non-physical} & \makecell[l]{protect} & \makecell[l]{backstage, co-located,\\public} & \makecell[l]{continuous}\\

\addlinespace[0.5em]

\makecell{\textbf{Museum Robot}\\Spelling Bee} & \makecell[l]{trained, designated} & \makecell[l]{physical\\non-physical} & \makecell[l]{teleoperate, maintain,\\troubleshoot} & \makecell[l]{co-located,\\frontstage, public} & \makecell[l]{planned, spontaneous,\\synchronous} \\

\midrule

\makecell{\textbf{Hospital Robot}} & \makecell[l]{highly trained,\\designated individual} & \makecell[l]{non-physical} & \makecell[l]{maintain,\\troubleshoot} & \makecell[l]{remote, non-public} & \makecell[l]{asynchronous, triggered} \\

\addlinespace[-0.7em]

\bottomrule
\end{tabular}
\caption{The wrangler profiles in each of our vignettes, characterized by our typology.}
\Description{A table summarizing wrangling profiles across several robot vignettes (help desk, fundraising, shopworker, museum, and hospital). Each row specifies who performs wrangling (typically trained, designated individuals or teams), the type of intervention (physical or non-physical), the purpose (e.g., setup, troubleshooting, teleoperation, maintenance, safety), where it occurs (co-located or remote; public or non-public; frontstage or backstage), and when it occurs (e.g., synchronous, asynchronous, planned, spontaneous, or continuous), highlighting variation in roles, contexts, and activities across scenarios.}
\label{table:vignettes}
\end{table*}
\normalsize

\subsection{Vignette: \textit{Help Desk Robot}}

\begin{figure}
    \centering
    \includegraphics[width=\columnwidth]{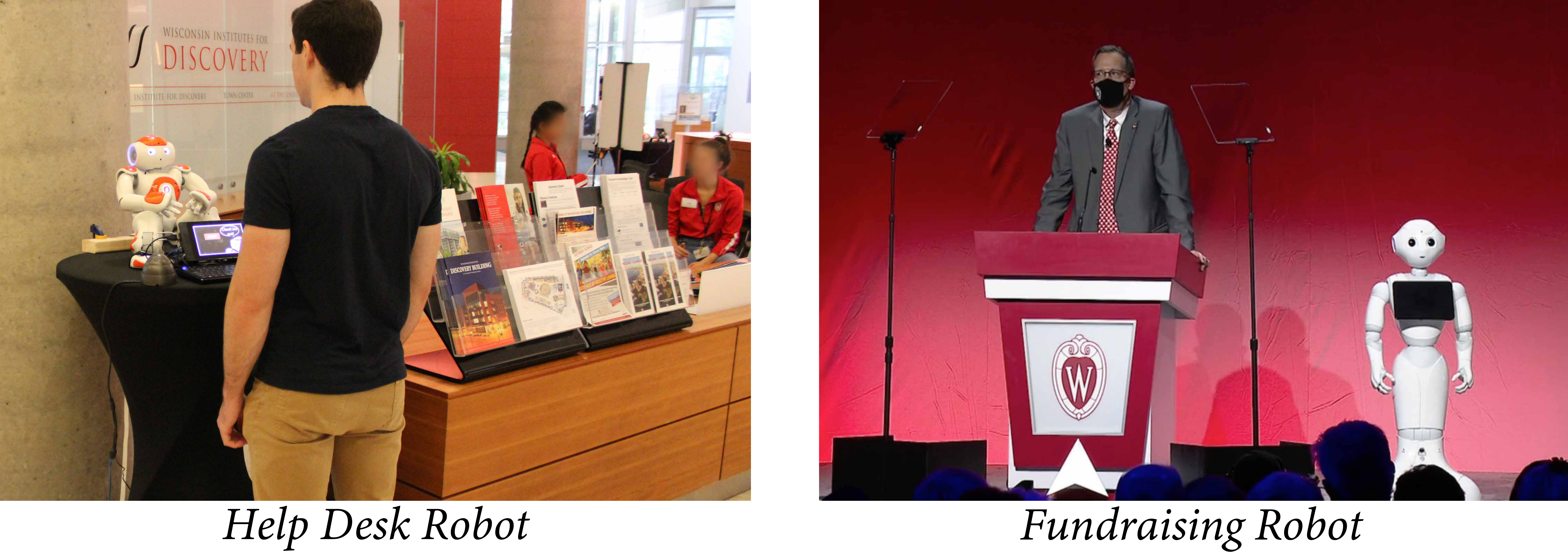}
    \caption{(Left) The physical setup of the \textit{help desk robot}, including the front desk and staff to the right of the robot. (Right) A teleoperated Pepper during a live fundraising ceremony, in which the robot’s stage placement, audience-facing orientation, and lack of visible operator cues contributed to a failure-intolerant wrangling scenario.\protect\footnotemark}
    \Description{Two side-by-side photos illustrating robot deployment contexts. Left: a small humanoid help desk robot on a table in a lobby or reception area, facing a standing person, with brochures and staff visible behind a front desk. Right: a humanoid Pepper robot on stage beside a podium during a formal fundraising event, oriented toward an audience, with a speaker at the podium and no visible operator. The images contrast a staffed, visible support setting with a public, audience-facing scenario where robot operation is less transparent and more failure-sensitive.}
    \label{fig:helpdesk}
\end{figure}
\footnotetext{Fundraising Robot Image \copyright 2026 Wisconsin Foundation \& Alumni Association.}

This vignette describes the first author's experience conducting a public deployment of two Nao robots \cite{gouaillier2009mechatronic} at the front desk of a university building in 2019 \cite{porfirio2020transforming}. \bl{}This vignette illuminates how \textit{low perceived risk} shapes a wrangler's decision to intervene over troubleshooting\bk{}.

\paragraph{\textbf{Context.}}Depicted in Figure \ref{fig:helpdesk} (left), the purpose of the deployments was to collect data on conversational interactions with bystanders via spoken language about parking and directions within the building. The deployments were limited to business hours for a time period of approximately one month. In conversing with bystanders, the robot moved its head and arms while talking in order to simulate human-like eye gaze and gestures. Other stakeholders included other bystanders in the building, workers at the front desk, and building IT. The wranglers consisted of the author and a team of undergraduate students who hovered behind the scenes in the vicinity of the robot. \bl{}The deployments were low risk, where failures amounted only to discarded data\bk{}.

%

\paragraph{\textbf{Wrangling Scenario: \textit{Intervention}.}} After a few weeks of deployment in public, a reoccurring, though rare defect emerged with the software that controlled the robot's gesture module. When the bug occurred, the conversation would freeze but the robot would continue to move its arms indefinitely until a wrangler arrived to physically reset the robot. Wranglers in the vicinity of the robot were usually able to provide a quick reset.

On one particular occasion, due to scheduling constraints, the primary wrangler stepped away from the robot to attend a meeting, and no other wranglers were available at the time. At an unknown point during the day after several hours of no wrangler supervision, the defect occurred. The wrangler arrived on the scene to a frozen robot with its joints overheating. The wrangler physically reset the robot, let its joints cool down, rebooted it, and allowed it to resume bystander interactions. Because the wrangler determined that \bl{}the cost of re-deploying the robot without a fix was lower than temporarily decommissioning the robot to fix the issue\bk{}, this scenario first involves \textit{intervention} rather than troubleshooting. Only after several more days of looking at sensor logs and debugging were the wranglers able to diagnose and patch the issue.

\subsection{Vignette: \textit{Fundraising Robot}}

This vignette describes the same author's experience controlling a Pepper robot \cite{softbankrobotics-pepper-productpage-2026} on stage at a university fundraising campaign that was simultaneously held in person and broadcasted online in the early 2020s. \bl{}This vignette shows how higher perceived risk can lead to extensive planning and affect teaming decisions to assist with continuous monitoring of the robot.\bk{}

\paragraph{\textbf{Context.}} This event was tightly controlled by the event organizers and the video production team responsible for broadcasting the event. Depicted in Figure \ref{fig:helpdesk} (right), robot engaged in a scripted interaction with a university speaker during the fundraising campaign. As the speaker spoke, the wrangler teleoperated the robot onto the stage, situated the robot next to the speaker, and briefly interacted with the speaker via dialogue. While speaking, the robot emitted non-verbal social cues such as head movements and arm gestures. Stakeholders included an audience of donors, the speaker, the event organizers, and the video production team. The wranglers included two graduate students---the author and another individual with robotics experience. A secondary wrangler assisted the primary wrangler by continuously monitoring the interaction.

\paragraph{\textbf{Wrangling Scenario: \textit{Preparation}.}} Being, failure-intolerant, this event was tightly controlled with a substantial amount of preparation needed by the wranglers. The robot's social cues in particular demanded immense temporal precision in order to align with the speaker's speech, and were thus highly rehearsed by the wranglers. The wranglers additionally needed to coordinate with the event organizers to ensure that the robot would only be on stage for a short period of time (as opposed to the entire speech) in order to prevent the robot's joints from overheating, which would cause a disruptive message in the middle of the speech.

In raising the stakes even further, the stage upon which the interaction occurred was raised approximately 1 meter above ground, leading to concerns that the robot would accidentally roll off the stage if it lost its wireless connection. To address these concerns and ensure the safety of both the robot and the audience, before the event, the wranglers developed and rigorously tested a failsafe mechanism that consisted of a twofold solution. First, the wranglers requested that the edge of the stage be augmented with a raised lip, which could easily be detected by the robot. Second, the wranglers added software that caused a full stop to the robot's base motion should a cliff be detected. This software required precise tuning the robot's cliff-detection sensor to prevent false positives from causing abrupt stops and false negatives to prevent the robot from rolling over the raised lip, off the stage, and into the crowd.

The teleoperation itself consisted of only four minutes and thirty seconds of the wrangler's time. By contrast, rehearsing social cues, coordinating with the event team, and developing the failsafe took approximately ten hours. To the wranglers' relief, the failsafe was not needed, and the event concluded without issue.

\subsection{Vignette: \textit{Shopworker Robot}}

This vignette pertains to the fourth and fifth authors' experiences deploying a human-like robot as an autonomous shop worker in a hat store in Japan, and highlights the different kinds of training and expertise that are often essential in the same wrangling task\bk{}.

\begin{figure}
    \centering
    \includegraphics[width=\columnwidth]{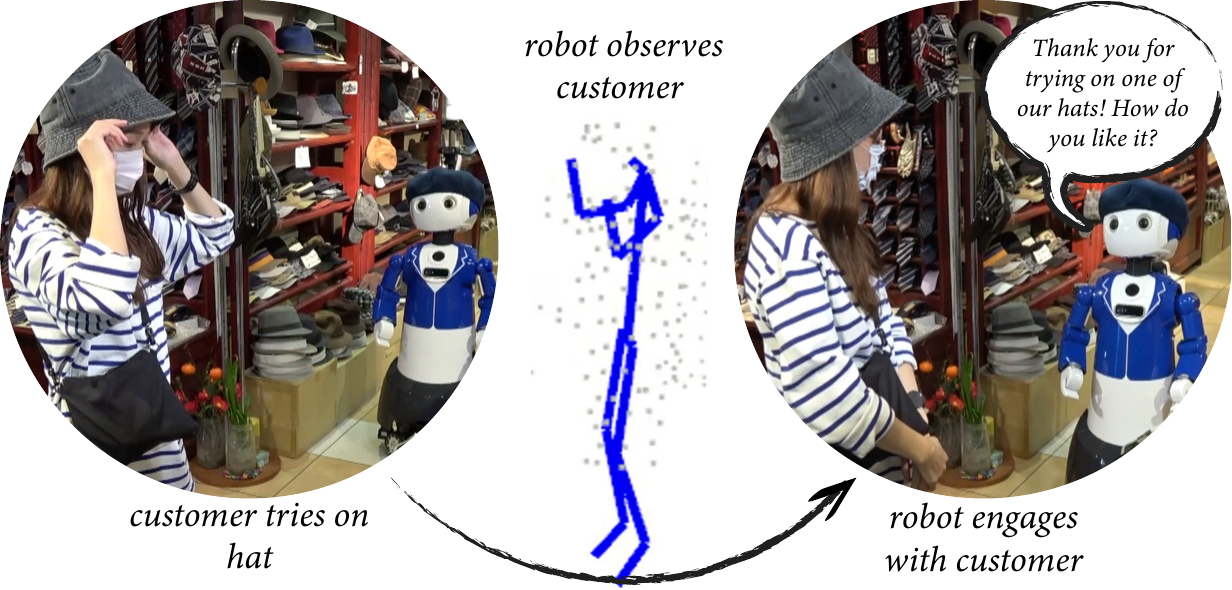}
    \caption{The shopworker robot detects the customer's action and starts to encourage them.}
    \Description{A three-part illustration showing a shopworker robot interacting with a customer in a retail setting. Left: a customer tries on a hat while standing near shelves of merchandise, with a small humanoid robot nearby. Center: a simplified skeletal visualization indicates the robot detecting the customer’s pose or action. Right: the robot faces the customer and engages them, with a speech bubble saying, “Thank you for trying on one of our hats! How do you like it?” The sequence illustrates the robot observing, interpreting, and responding to the customer’s behavior.}
    \label{fig:shopworker}
\end{figure}

\paragraph{\textbf{Context.}} The robot initially provided advertisements at the shop entrance to attract pedestrians, and once they entered the shop, approached them and encouraged them to try on a hat by detecting the customer's action (Figure \ref{fig:shopworker}).
The wranglers publicly advertised the field study but did not give any instruction for these visitors before interacting with the robot.

\paragraph{\textbf{Wrangling Scenario: Prototyping.}} 
This robot was developed iteratively by team members with different perspectives: HRI experts, engineers, and assistant staff who are not experts in robotics or human-robot interaction.
The team discussed the initial design of the robot, namely interaction strategies, robot capabilities, and appearance.
Once a prototype of the system was ready, the team tested the prototype at the shop to get feedback from within the team itself and a shop manager.
The purpose of the testing was to improve interaction design and debug.
During this testing, the team sometimes tested the robot with actual customers in the same setting of the eventual actual development. 
One clear example of changes to the interaction design came from shop manager feedback.
In the first design, our robot was confined to within the shop, but the shop manager requested that it invite pedestrians from outside of the shop.
Thus, the team changed the robot to move to the front of the shop entrance when there were no customers in the shop and repeated this iterative testing cycle to improve the robot's interaction design.

\paragraph{\textbf{Wrangling Scenario: \textit{Deployment}.}} 
During the field trial, two staff were assigned to the robot: one local safety staff (usually a non-expert of robotics and human-robot interaction) and one remote expert staff (the expert of this robot system).
The team trained local safety staff to handle simple and typical problems.
Local staff handled simple troubleshooting and ensure safety (\ie{} stop the robot when a child ran to it).
The team prepared a manual of typical error cases for the local safety staff, such as for hardware errors and navigational errors, as this robot was sometimes placed in narrow spaces and needed to be remotely maneuvered via a controller.

However, the team sometimes faced complicated robot issues that the local staff could not solve according to their manual, upon which a remote expert connected to the robot system to resolve them.
The reasons for these problems are wide-ranging.
Sometimes, they were due to a rare error that was not described in the manual, such as the disconnection of a sensor cable.
Other times issues were caused by unexpected human behavior or bugs.
In this case, even for the expert, it was difficult to solve problems in a timely manner.
Thus, the expert showed a temporary solution to the local staff while the expert laboriously sifted through the robot's sensor logs and internal state to form a hypothesis about unexpected behaviors.
After the analysis, the expert either updated the robot system or discussed with the team about further iterations to the robot's interaction design  for future updates.



\subsection{Vignette: \textit{Museum Robot}}

In the late 2010s, the Hirshhorn museum in Washington, D.C., deployed the Pepper robot in several different settings. Here, we present a collection of vignettes that describe the second author's experience as a \bl{}firsthand robot wrangler and in observing other wranglers secondhand. These vignettes highlight additional contextual dimensions that pertain to wrangling, specifically the physical capabilities of the robot, its physical affordances, and its autonomy. They also highlight an additional dimension of the wrangler's training and experience---their \textit{risk aversion}---leading them to increase their monitoring over the robot and their control over the robot\bk{}.



\paragraph{\textbf{Context}}

The robot was used in several different ways at the museum. In one instance, the museum's makerspace ran a summer program for teenagers where Pepper was utilized as a teaching tool. The robot’s software environment, \textit{Choregraphe} \cite{pot2009choregraphe}, alongside its many sensors, provided an engaging and accessible way to teach youth various forms of technology. Stakeholders included the class of teenagers learning how to program. The wrangler consisted of a single individual---the author---whose primary job it was to plan, setup, and takedown learning sessions, teach students, and troubleshoot glitches on the robot when necessary.

In other instances, the robot was deployed in public engagements within the museum. In one such deployment, Pepper acted as an (artificially) live model, responding to users' voice requests to imitate various artwork in the museum, while displaying information about the artwork on its tablet. Other deployments involved placing the robot near the front entrance to the museum, where the robot repeated curated information about the museum and its current exhibits, and as the host of the \textit{Smithsonian Spelling Bee}. Stakeholders in these deployments involved bystanders and museum security, who acted as wranglers themselves to ensure the safety of the robot. The primary wranglers included several designated engagement coordinators with varying degrees of training.

\paragraph{\textbf{Wrangling Scenario: Teenager Education}} For educating teens on robot programming, the wrangler conducted a substantial amount of asynchronous lesson planning to ensure that learning objectives were met. Although hardware and software glitches inevitably occurred during lessons, these were embraced as valuable learning experiences. In these situations, the wrangler simply demonstrated troubleshooting strategies to the class in real time.

\paragraph{\textbf{Wrangling Scenario: Affordance Mismatch}} Wrangling exhibited different characteristics outside of the educational setting. For our second scenario, while deployed in public settings, \bl{}in order to maintain safety of the robot and bystanders around it, the robot was tightly controlled (\ie{} not allowed to locomote) placed under constant monitoring by engagement coordinators and security staff\color{black}. Visitors often approached the robot and attempted to physically interact with it out of curiosity. The robot's hands generated substantial interest, implying a greater-than-actual ability to grasp objects and interact with visitors via touch. This led to continued user interaction with the robot's hands, eventually breaking them. 

\paragraph{\textbf{Wrangling Scenario: Spelling Bee}} During Pepper's time at the museum, the wrangler was asked to prepare the robot to participate in the annual \textit{Smithsonian Spelling Bee} by responding to spelling prompts with several humorous lines and lifelike animations. On the day of the event, the audience watched contestants, the host, and Pepper participate onstage. Because Pepper's voice recognition was insufficient for hearing prompts, the wrangler manually cued animations via Choregraphe. Despite the wrangler rigorously pre-testing the robot's behaviors, Pepper glitched in the middle of the event and was unable to respond when cued. After an unsuccessful attempt to reset the robot, it was wheeled backstage for later troubleshooting, upon which the wrangler, who is adept in live comedy, attempted to diffuse the awkwardness of the situation.

\subsection{Imagined Vignette: \textit{Hospital Robot}}

Hospitals are ideal environments to reflect on human interaction with technology and the future deployment of robots. Here, we present a forward-looking vignette that utilizes the third author's expertise to envision robot wrangling in hospitals. \bl{}This vignette highlights an additional dimension of wrangling not captured in our typology---\textit{institutional structure}\bk{}.

\paragraph{\textbf{Context}} Hospitals are technologically complex environments with groups of engineers and technicians required on-site to ensure the functionality of safety-critical equipment. This highly trained stakeholder, the \textit{bioengineer}, is in charge of machine setup, troubleshooting, maintenance, and repair. If they cannot handle repair, they will coordinate with the machines' vendors for maintenance. 
In addition to technical challenges, bioengineers must navigate the relationship between robots and humans. 
Although robots can help address patients’ needs and even serve as companions during hospitalization, patients might feel more comfortable working with a real human being, especially when clinical care is very personal and life-critical. 
Staff may also feel they might need to take care of robots, which are poised to negatively affect them if not properly integrated into their workflows. 
As wranglers, bioengineers will need to address these human concerns.


\paragraph{\textbf{Wrangling Scenario: Navigating Institutional Structure}} Consider a nurse at a hospital who wishes to use a robot to deliver medication to patients. The nurse is busy with another patient and does not have any free hands. They attempt to verbally instruct the robot, but it struggles to understand their speech. The nurse raises the issue with the robot wrangler, a highly trained bioengineer, who begins monitoring for other occurrences of the issue. Recognizing that they must navigate institutional constraints while keeping patients' lives at the forefront, the wrangler has two options---limit the robot's voice recognition capabilities to keyword recognition (a quick fix) or re-parameterize and re-train the robot's voice recognition module (a more robust, though difficult fix). The bioengineer chooses the more robust option, which requires weeks of testing and validation. During this testing, the nurses raise the problem with hospital administration, who, rather than allowing the wrangler time to test and validate the technical fix, change the medication-delivery schedule to align robot usage with lighter nurse workflows. This inadvertently causes further tension with the nurses, who had become accustomed to the old schedule.

\bl{}
\subsection{Themes}

Our vignettes surface several themes, which we present below.

\paragraph{\textbf{Theme 1:} risk is tied to wrangling decisions and outcomes. } In our brainstorming sessions, \textit{risk} emerged as a dimension that shapes wrangling decisions. In both the \textit{help desk robot} and the \textit{spelling bee} scenario of the \textit{museum robot}, the low amount of risk associated with the consequences of failure led the wrangler to intervene spontaneously rather than extensively plan for contingencies asynchronously. Conversely, the high-stakes environment present in the \textit{fundraising robot} led the wrangler to asynchronously and meticulously prepare the teleoperation session in order to completely mitigate all failure risk. In our experience as wranglers, risk is tied to whether wrangling decisions are planned beforehand or spontaneously made in the moment.

Related to risk \textit{level}, risk \textit{tolerance} was discussed as affecting the second author's wrangling decisions in the \textit{museum robot} vignette. Specifically, low risk tolerance led the wrangler to maintain tight control over the robot via pre-scripted interactions and have the robot remain stationary rather than locomote. Low risk tolerance was additionally attributed to a perceived need to continuously monitor the robot. We therefore observe a tradeoff between wrangler risk tolerance and the degree to which the robot is continuously monitored and used overall.

\paragraph{\textbf{Theme 2:} institutional structure constrains wrangling decisions. } \textit{Institutional structure} similarly emerged from extensive discussions with the third author about the \textit{hospital robot} as another mechanism for explaining wrangler decisions. Wranglers must navigate institutional constraints, including how their decisions for \textit{what} wrangling activity to pursue and \textit{when} to pursue it may conflict or synergize with other institutional stakeholders. We therefore observe tensions between \textit{when} and \textit{how} wrangling occurs and institutional structure, including which conflicting interests the wrangler may need to prioritize (\eg{} administration versus nurses). 

\paragraph{\textbf{Theme 3:} wrangling is both technical and sociocultural. } The \textit{shopworker robot} vignette highlights how issues that require a wrangler's attention can either be technical (\eg{} in the \textit{deployment}) scenario or sociocultural (\eg{} in the \textit{prototyping}) scenario. Both types of issues are similarly present in the \textit{museum robot}, in which a mismatch between a social affordance (the robot's hands) and the inherent curiosity of museum goers led to physical breakage. 

This leads us to a subtheme---\textbf{sociocultural breakdowns can arise from bystander perceptions of robot affordances.} 
In the case of the \textit{museum robot}, the sociocultural breakdown led to a need for a technical fix. We also discussed how this failure would be unlikely to occur in the \textit{hospital robot} vignette due to the different social norms of the hospital environment, specifically how bystanders may be less likely to approach robots with such curiosity in hospitals.

\paragraph{\textbf{Theme 4:} wrangling is a temporal, sometimes retrospective phenomenon. } It is evident from all of our scenarios that wrangling involves a temporal evolution of activity, involving different stakeholders in different places over time. 
In particular, the \textit{teen education} scenario in the \textit{museum robot} vignette demonstrates a cycle between asynchronous planning and synchronous troubleshooting activity. The \textit{shopworker robot prototyping} scenario  further exemplifies the temporal nature of wrangling, where the wrangling team tuned the robot's social interactions over multiple iterations of interaction design with the shop owner and customers.

In the \textit{shopworker} and \textit{help desk} vignettes, wrangling eventually shifted to a \textit{retrospective} activity, highlighting how wranglers sometimes need to sift through log data to hypothesize the causes to observed issues. In our discussions as wranglers, we realized the importance of hypothesis formation being a primarily wrangler-driven, rather than automated process, especially due to the sociocultural nature of may of these issues. Though automated assistance for hypothesis formation and testing can be a welcome addition to a wrangler's toolkit, we stressed the importance of such assistance not artificially biasing the wrangler towards different conclusions.

\bk{}

\section{Discussion}

\bl{}We present several prescriptive design implications from the synthesis of our scoping review and vignettes. Inspired by the technology support afforded to other stakeholders---authoring tools for end users, teleoperation tools for operators, and prototyping tools for designers---our first set of implications pertain to technology support for wranglers, recognizing that a catch-all ``wrangling tool'' is infeasible due to the diversity of the practice. Our second set of implications pertains to institutional practices, methods for improving the epistemic inquiry of wrangling before robots are embedded in institutions, and institutional policy. 

\subsection{Implications for Technology Design}



\textbf{Wrangling tools should calculate risks to help wranglers navigate tradeoffs.}
Our vignettes reveal a fundamental tradeoff between \textit{intervention effort} and \textit{pre-planning effort} mediated by contextual risk level and personal risk tolerance. In the \textit{fundraising robot} scenario, extensive preparation and rehearsal reduced the need for live intervention, shifting effort upstream to anticipate and mitigate deployment risks. By contrast, the \textit{museum robot} required continuous monitoring and readiness to intervene, as unanticipated bystander interactions created persistent uncertainty that had not been fully accounted for during preparation. Tools should therefore support wranglers in navigating this tradeoff by making the costs and benefits of pre-planning versus real-time intervention visible, particularly through calculating and surfacing the risks associated with either option. 
Calculating risks necessitates that tools are equipped with predictive capabilities, including simulation and verification of robot behaviors under varied conditions, as well as mechanisms for encoding contingency plans.

\textbf{Wrangling tools should support team coordination.}
As evidenced by our typology and the \textit{shopworker robot} vignette, wrangling is frequently distributed across individuals with different backgrounds, expertise, roles, and temporal availability. The \textit{shopworker robot} illustrates configurations ranging from tightly coupled teams during prototyping to loosely coordinated actors during long-term deployment, where expert support may be required asynchronously. Tools should therefore facilitate coordination across expertise levels, including mechanisms for handing off issues, annotating incidents, and maintaining shared records of past interventions. This includes supporting both synchronous collaboration (\eg{} multiple wranglers jointly supervising a deployment) and asynchronous escalation (\eg{} frontline staff deferring complex issues to expert roboticists). Designing for coordination also entails making system state and history legible to diverse stakeholders, reducing reliance on tacit knowledge and ad hoc communication.

\textbf{Wrangling tools should support not just technical, but  \textit{sociocultural} alignment.}
Our lived experiences as wranglers revealed to us that wrangling is often not purely a technical phenomenon, but a practice of aligning user expectations and social norms with the robot's social behaviors and affordances. This is especially evident in the \textit{shopworker robot} and \textit{museum robot} vignettes. Wrangler support tools must therefore support this alignment. One approach employed by the first author in prior work is \textit{embodied design} for allowing individuals to rehearse interactions by ``stepping into the shoes of'' other stakeholders, explore social breakdowns, and iteratively refine robot behavior \cite{porfirio2019bodystorming, pelikan2023designing}. 
Such support is particularly critical for preparing for edge cases that may not be captured in formal specifications but emerge in situated deployments.

\textbf{Wrangling tools should support retrospective sensemaking for troubleshooting.}
Our vignettes expand our understanding of \textit{troubleshooting}, which becomes increasingly crucial when constant monitoring is impractical, such as in the \textit{help desk robot}. Tools should therefore aggregate multimodal data sources---including video, audio, and system logs---into temporally aligned, queryable representations, and then assist in \textit{hypothesis formation} by surfacing anomalies, suggesting plausible causal factors, and enabling wranglers to compare across similar prior incidents. Customizable visualizations and filtering mechanisms are critical to accommodate differences in expertise (\eg{} roboticists vs. frontline staff) and to support both synchronous and asynchronous troubleshooting workflows. However, in order to navigate the tensions surrounding AI-assistance in hypothesis formation, these interfaces must strategically (possibly even sparingly) use such assistance in order to avoid biasing wranglers towards certain causes behind an issue.

\subsection{Implications for Institutional and Methodological Practices}

While our prior implications emphasize technosolutionism, equally important is supporting wranglers indirectly through institutional and methodological practices. 

\textbf{Designers should use multilevel service blueprinting to explicitly map out wrangling labor, using our typology as a starting point, before embedding robots within distributed human ecologies.} Doing so will reframe HRI from a purely user-centric phenomenon to include the labor that is both visible invisible to the user and understand who performs this work, when it occurs, and how it interacts with other stakeholders. The outcome of service blueprinting would be to foreground and avoid cases such as \textit{patchwork} \cite{fox2023patchwork}, where undesignated personnel like the security staff of our \textit{museum robot} vignette fall into the role of wrangling by necessity, or bystanders in proximity to the robot are co-opted into an \textit{upkeep} role.

\textbf{Designers should adopt our typology as a formal documentation scaffold.} It is evident from our review that what researchers label as ``wrangling'' spans a wide and heterogeneous set of activities, roles, and contexts rather than a single coherent practice. Yet, when this diversity is collapsed into a single term, important forms of labor---particularly those that are backstage, intermittent, or distributed across multiple actors---risk becoming obscured or taken for granted, and wrangling decisions that are potentially useful to others remain hidden. Simply reporting that ``wrangling occurred'' offers little insight into what was actually done, by whom, under what conditions, or with what implications for system design and deployment. To address this, we advocate that designers who document their practices move beyond treating wrangling as a catch-all label and instead adopt our typology as a reporting scaffold. Papers that describe wrangling should explicitly characterizing who performs it, what wrangling actions are taken, where and when it occurs, and for what purpose. In doing so, researchers can make this labor visible, comparable and actionable, and ultimately enable more precise understanding and better support for the people who sustain real-world robot deployments.

We are also aware that labeling what is referred to as \textit{patchwork} and \textit{upkeep} in the related literature as ``wrangling'' risks glorifying the labor of individuals whose availability and proximity to the robot is taken advantage of. We therefore place special emphasis on the importance of clearly reporting the status of these individuals not simply as ``wranglers.''

\textbf{Designers can leverage participatory and co-design workshops to both surface and affect institutional wrangling policy.}
The \textit{hospital robot} vignette shows that wrangling is embedded within broader institutional contexts, where risks, policies, and third-party constraints shape what actions are possible or appropriate. Diverging wrangler and stakeholder constraints create a tension that the wrangler must navigate. Collaborative design, including both participatory design and co-design, presents an opportunity to shape institutions, either through policy or technology design, towards streamlining the communication of institutional knowledge to the front-line wranglers, and wrangler decision making back to the wider institution. Designers who work with institutions in a collaborative design capacity can use activities such as service blueprinting workshops and scenario-based walkthroughs of wrangling practices to ensure that wrangling concerns are explicitly surfaced, negotiated, and made accountable. 





\bk{}

\subsection{Limitations and Future Work}

Our work is not without limitations. First, while our scoping review captures a wide breadth of wrangling usage in the literature, it fails to capture the same level of depth as a \textit{systematic} review. As a result, we do not capture wrangling-adjacent terms like \textit{handling} \cite{woods2004envisioning} and \textit{patchwork} \cite{fox2023patchwork}. Our decision to focus on breadth rather than depth was practical---\textit{all} robotics involves robot wrangling to some extent, even if invisible or undiscussed in publications, calling into question the feasibility of a systematic review.

Second, our typology is a starting point. We acknowledge that our  focus with social and service robotics omits other fields, such as search and rescue, collaborative robot manipulators, and autonomous vehicles, to name a few examples. Thus, additional perspectives from these fields are needed to form a more complete picture of robot wrangling, in addition to other domains that we missed (hotels, warehouses, and household robots, to name a few examples). We therefore believe that the best first step to extending our typology is to speak with stakeholders in these domains. 


\bl{}
Lastly, several categories of wrangling remain woefully unexplored in the  literature. Despite the strong connections to concepts like \textit{patchwork} \cite{fox2023patchwork}, \textit{upkeep} \cite{dobrosovestnova2025beyond, tornberg2021investigating}, and \textit{helpers} \cite{dobrosovestnova2025beyond, fallatah2020would, yu2024encouraging}, there are many opportunities for further design inquiry that include participatory co-design with \textit{undesignated} wranglers, for which our scoping review captures only two works \cite{bjorling2022designing, lee2025minding}, and \textit{untrained} wranglers, for which our scoping review captures four works, though all of which being in an educational context \cite{hunt2025children, 
silvis2022technical, 
benedict2019application, 
anderson2024integrated}. We remain interested in the activities of individuals like the wranglers in the \textit{museum robot} vignette, who informally fell into a designated wrangling role, in addition to the security staff who continuously monitored the robot as expensive research equipment.
\bk{}

\section{Conclusion}

\bl{}Researchers are \color{black} increasingly referencing \textit{robot wranglers} as key stakeholders in human-robot interaction. This paper strives to understand robot wrangling in order to design support for this new stakeholder. To this end, we conduct a scoping review of research literature that describes robot wrangling, and contextualize this review with our own lived and imagined experiences as wranglers. \bl{}We then provide several implications for how interactive systems should be designed to support this stakeholder\color{black}. 

\begin{acks}
This work was supported by George Mason University; the University of Maryland, College Park; and the Center for Nursing Research, Education, and Practice at Houston Methodist Hospital; and the JST Moonshot R\&D Program (Grant No. JPMJMS2011, Japan) and JSPS KAKENHI (Grant No. 24H00722, Japan).
\end{acks}

\bibliographystyle{ACM-Reference-Format}
\bibliography{references, litreview}

\end{document}